\documentclass[conference]{IEEEtran}
\IEEEoverridecommandlockouts
\usepackage{subfigure} 
\usepackage{cite}
\usepackage{amsmath,amssymb,amsfonts}
\usepackage{algorithmic}
\usepackage{graphicx}
\usepackage{textcomp}
\usepackage{xcolor}
\usepackage{booktabs} 
\def\BibTeX{{\rm B\kern-.05em{\sc i\kern-.025em b}\kern-.08em
    T\kern-.1667em\lower.7ex\hbox{E}\kern-.125emX}}
\begin{document}

\title{
Enhancing Interpretability of AR-SSVEP-Based Motor Intention Recognition via CNN-BiLSTM and SHAP Analysis on EEG Data
}

\author{
\IEEEauthorblockN{Lin Yang}
\IEEEauthorblockA{
\textit{School of Control} \\
\textit{Science and Engineering} \\
\textit{Shandong University} \\
Jinan, Shandong \\
202334983@mail.sdu.edu.cn
}
\and
\IEEEauthorblockN{Xiang Li}
\IEEEauthorblockA{
\textit{School of Control} \\
\textit{Science and Engineering} \\
\textit{Shandong University} \\
Jinan, Shandong \\
202120648@mail.sdu.edu.cn
}

\and
\IEEEauthorblockN{Xin Ma*}
\IEEEauthorblockA{
\textit{School of Control} \\
\textit{Science and Engineering} \\
\textit{Shandong University} \\
Jinan, Shandong \\
maxin@sdu.edu.cn
}
\and
\IEEEauthorblockN{Xinxin Zhao*}
\IEEEauthorblockA{
\textit{ Shandong Inspur } \\
\textit{Science Research Institute Co., Ltd.} \\
Jinan, Shandong \\
zhaoxxjs@inspur.com
}
}

\maketitle

\begin{abstract}
Patients with motor dysfunction show low subjective engagement in rehabilitation training. Traditional SSVEP-based brain-computer interface (BCI) systems rely heavily on external visual stimulus equipment, limiting their practicality in real-world settings. This study proposes an augmented reality steady-state visually evoked potential (AR-SSVEP) system to address the lack of patient initiative and the high workload on therapists. Firstly, we design four HoloLens 2-based EEG classes and collect EEG data from seven healthy subjects for analysis. Secondly, we build upon the conventional CNN-BiLSTM architecture by integrating a multi-head attention mechanism (MACNN-BiLSTM). We extract ten temporal-spectral EEG features and feed them into a CNN to learn high-level representations. Then, we use BiLSTM to model sequential dependencies and apply a multi-head attention mechanism to highlight motor-intention-related patterns. Finally, the SHAP (SHapley Additive exPlanations) method is applied to visualize EEG feature contributions to the neural network's decision-making process, enhancing the model's interpretability. These findings enhance real-time motor intention recognition and support recovery in patients with motor impairments.

\end{abstract}


\renewcommand{\thefootnote}{} 
\footnotetext{
National Key Research and Development Program Project under Grant 2023YFB4706104.
Key R\&D Program of Shandong Province (Major Science and Technology Innovation Project) under Grant 2024CXGC010603.
}
\renewcommand{\thefootnote}{\arabic{footnote}} 

\section{Introduction}

BCI technology offers an alternative communication pathway for individuals with motor impairments by enabling interaction without relying on muscular activity. It holds promising applications in fields such as neurorehabilitation and assistive device control~\cite{b1, b2}. Motor dysfunction is a common consequence of various neurological conditions, including stroke, spinal cord injury, amyotrophic lateral sclerosis (ALS), cerebral palsy, and Parkinson’s disease~\cite{b3,b4,b5}. These conditions may lead to partial or complete paralysis, significantly affecting the ability to perform daily activities and thus reducing the quality of life. In some cases, patients may also suffer from disorders of consciousness or aphasia, further restricting their ability to communicate with the external environment.

In recent years, BCI technologies have advanced significantly in rehabilitation engineering and have been widely applied to the control of assistive devices~\cite{b6}. To improve patient engagement and rehabilitation outcomes, various BCI-driven exoskeleton control strategies have been proposed~\cite{b7}. For example, Missiroli et al. developed a motor imagery (MI)-based system enabling task activation via imagined limb movements~\cite{b8}, while others have embedded SSVEP into exoskeleton control for precise intention decoding~\cite{b9}. P300-based approaches have also been explored for intention recognition and feedback~\cite{b10}. Among these, SSVEP-BCIs have gained prominence due to minimal training requirements, high signal-to-noise ratios, and support for multiple commands. These systems have shown superior classification accuracy and information transfer rate (ITR), making them a leading approach in BCI-driven rehabilitation~\cite{b11}.

However, current SSVEP-based BCIs still face several practical limitations. Traditional SSVEP-BCI systems typically rely on static displays or LED arrays for visual stimulation, requiring users to maintain fixation on specific regions, which greatly restricts their applicability in more naturalistic and immersive interaction scenarios~\cite{b12,b13}. Such rigid visual setups can also contribute to user fatigue and reduce long-term system usability, especially during prolonged rehabilitation sessions. Furthermore, although deep learning techniques have demonstrated impressive performance in EEG decoding tasks, their inherent "black-box" nature presents significant challenges in clinical applications, where transparency and explainability of model behavior are crucial~\cite{b14}. The lack of interpretability impedes clinicians' understanding of the system’s decisions.

To address these limitations, this study proposes the integration of AR into the SSVEP task design, enabling flexible and efficient visual interaction in natural environments~\cite{b16}. Additionally, we adopt the SHAP method to interpret the deep neural network's decision process, allowing for quantitative analysis of feature and channel contributions~\cite{b17}. This approach enhances system interactivity, interpretability, and clinical applicability, offering a new direction for practical rehabilitation-oriented BCI systems.

\section{Materials and methods}

\begin{figure}[b]
\centerline{\includegraphics[width=1\linewidth]{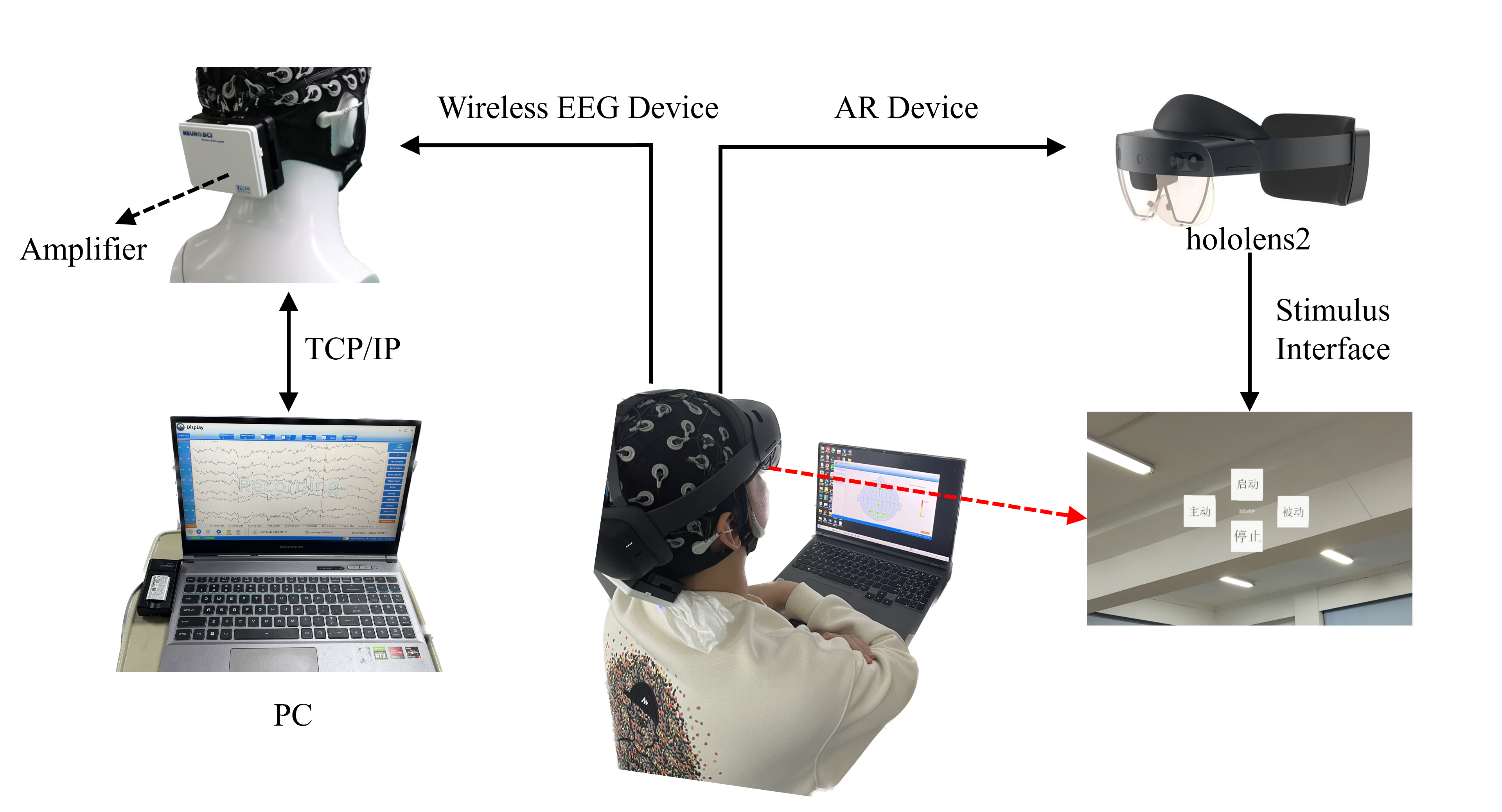}}
\caption{AR-BCI motor intention recognition system.}
\label{fig1}
\end{figure}

\subsection{AR-BCI Platform}\
In this study, an augmented reality-based brain–computer interface (AR-BCI) platform was developed by integrating the NeuroSci wireless EEG acquisition system with the Microsoft HoloLens2 head-mounted AR device, as illustrated in Fig.~\ref{fig1}. The NeuroSci system consists of an EEG cap, signal amplifier, synchronizer, and a host computer, and supports wireless acquisition of up to 64 EEG channels. It incorporates advanced technologies such as high-sensitivity sensor arrays, low-noise signal amplification, and high-throughput real-time data transmission, thereby ensuring high-fidelity signal acquisition and stable data communication. The system features ultra-low input noise and an exceptionally high common-mode rejection ratio, with a maximum per-channel sampling rate of 1000 Hz.

The HoloLens2 serves as a fully integrated spatial computing platform, equipped with a high-performance processing unit and a suite of embedded sensors capable of real-time tracking of head orientation and spatial positioning. It facilitates the presentation of visual stimuli in holographic form and supports intuitive, naturalistic human–computer interactions. In this study, the HoloLens2 was employed to deliver SSVEP stimuli. Meanwhile, synchronization with the NeuroSci system was achieved via a dedicated synchronizer to ensure precise temporal alignment between stimulus presentation and EEG signal acquisition.

By virtue of its wireless, wearable, and interactive design, the proposed AR-BCI platform offers a robust and flexible solution for immersive neural modulation and cognitive neuroscience experiments, providing strong technical support for the development of next-generation neurorehabilitation and brain–computer interaction systems.

\subsection{Subjects}\
Seven subjects(five males and two females, aged between 24 and 30 years) were recruited for this study. All subjects had normal or corrected-to-normal vision, normal stereoscopic perception, and no known history of neurological disorders or injuries. To ensure consistency, all subjects completed a task-specific training session before the experiment, even though some had prior exposure to similar SSVEP experiments. All experimental procedures were carried out after written informed consent was obtained from each subject. During the experiment, subjects were instructed to follow standardized task instructions and to remain highly focused. Data from all seven subjects were included in the final analysis.

\subsection{EEG recordings}\
The experiment was conducted in an electromagnetically shielded room to minimize interference from external noise sources. Continuous EEG signals were recorded using the NeuroSci acquisition system from eight electrodes (Oz, O1, O2, POz, PO3, PO4, PO5, and PO6) strategically positioned over the parietal and occipital regions, which are known to be highly responsive to visual stimuli. The visual stimuli were presented via the HoloLens 2 augmented reality headset. A reference electrode was placed at the central scalp region (Cz), while the ground electrode was positioned on the forehead (Fpz). Electrode impedances were consistently maintained below 10~k$\Omega$ to ensure signal quality and reliability. To enable precise temporal alignment between stimulus events and EEG data, event markers were employed for synchronization. These markers were transmitted from the stimulus control computer to the EEG synchronization module through a parallel port interface, thereby ensuring accurate event-related data extraction during subsequent analysis.

\begin{figure}[b]
\centerline{\includegraphics[width=0.8\linewidth]{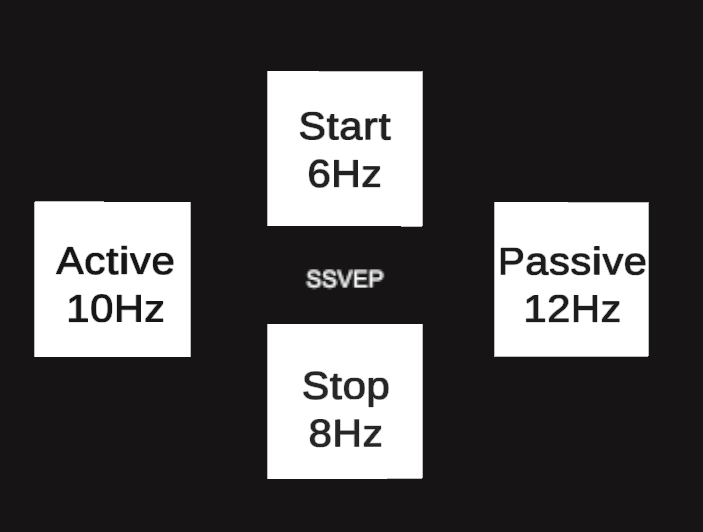}}
\caption{Visual stimulation interface.}
\label{fig2}
\end{figure}

\subsection{Stimuli and Experiment Design}\
SSVEPs are steady neural responses elicited by periodic visual stimuli, characterized by an enhancement of the stimulus frequency and its harmonics in EEG signals.

\begin{figure}[htbp]
\centerline{\includegraphics[width=1\linewidth]{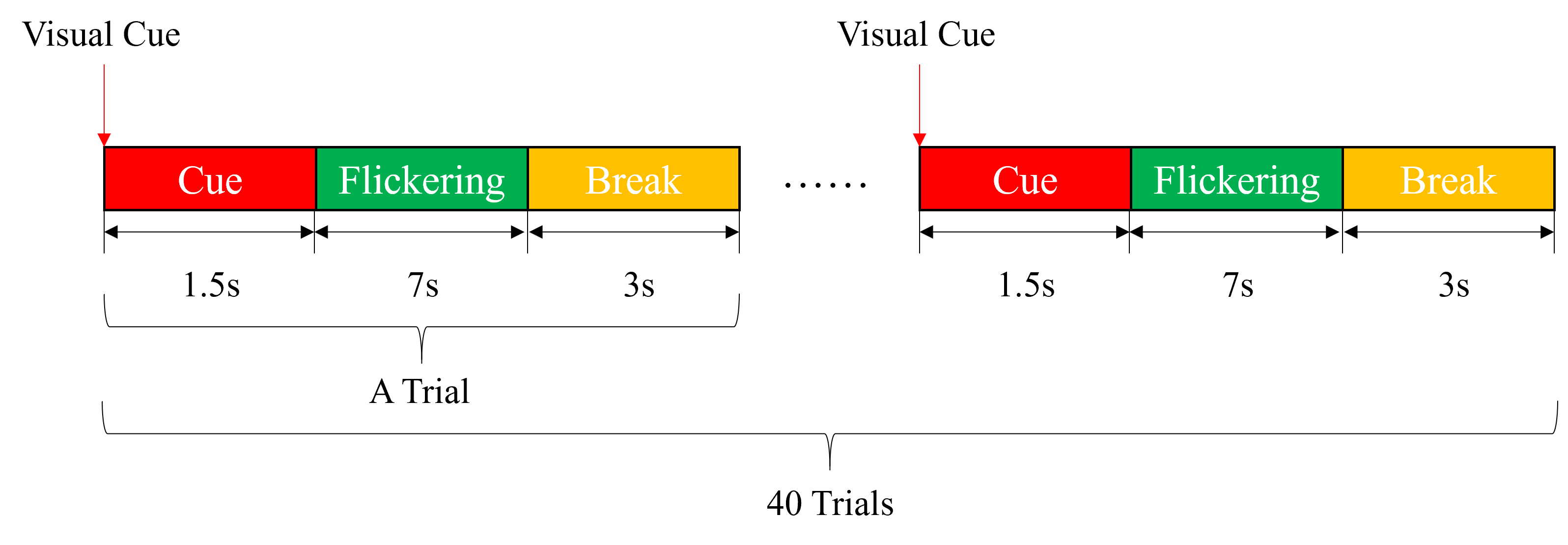}}
\caption{Process of SSVEP stimulus test.}
\label{fig3}
\end{figure}

The visual stimulation interface was designed using Unity 3D, as illustrated in Fig.~\ref{fig2}. The interface consisted of a $2\times2$ matrix layout, with each flickering block corresponding to a specific control command: Start (6~Hz), Stop (8~Hz), Active (10~Hz), and Passive (12~Hz), respectively labeled as 0, 1, 2, and 3. Each command was mapped to a distinct locomotion mode of the lower-limb exoskeleton, enabling intention recognition and interactive control via the BCI. Considering that low-frequency SSVEP stimuli typically exhibit higher signal-to-noise ratio (SNR), low-frequency visual flickers were selected in this study~\cite{b18}. To reduce visual fatigue while maintaining sufficient evoked responses, all stimuli were rendered in white~\cite{b19}. As shown in Fig.~\ref{fig3}, each trial began with a 1.5-second red cue to attract the subject’s attention. This was followed by a 7-second flickering phase according to predefined frequency and phase parameters, allowing for adequate EEG signal accumulation. After each stimulation, a 3-second rest period was introduced to alleviate visual fatigue and reduce cognitive load, thereby enhancing both data quality and experiment sustainability. Each subject completed a total of 40 trials to collect sufficient neural response data. On average, each stimulus class (Start, Stop, Active, Passive) was presented in 10 trials, ensuring balanced data distribution. To improve experimental flexibility and subject comfort, the system allowed data acquisition to be paused at any point upon request and resumed later, ensuring the completeness and validity of the dataset.

\subsection{EEG Processing}\
To mitigate noise interference, the system implements a series of preprocessing filters, including a low-pass filter with a cutoff frequency of 25 Hz, a high-pass filter with a cutoff frequency of 4 Hz. Considering the visual delay after triggering the visual stimulus, only the EEG data of the middle 4 seconds were selected for the experiment.
\subsection{Classification Methods}\
\paragraph{Canonical Correlation Analysis} 
Canonical Correlation Analysis (CCA) is a widely used method for classifying steady-state visual evoked potentials. It measures the linear correlation between two multivariate signals: the recorded EEG signal $\mathbf{X} \in \mathbb{R}^{N \times T}$ and the reference signal $\mathbf{Y} \in \mathbb{R}^{M \times T}$ constructed from sine-cosine components of the stimulus frequency and its harmonics~\cite{b20}. Here, $N$ is the number of EEG channels, $M$ is the number of reference components, and $T$ is the number of time points.

CCA seeks linear projections $\mathbf{w}_x$ and $\mathbf{w}_y$ such that the correlation between the projected signals $\mathbf{w}_x^\top \mathbf{X}$ and $\mathbf{w}_y^\top \mathbf{Y}$ is maximized:

\begin{equation}
\rho = \max_{\mathbf{w}_x, \mathbf{w}_y} \frac{\mathbf{w}_x^\top \mathbf{X} \mathbf{Y}^\top \mathbf{w}_y}{\sqrt{(\mathbf{w}_x^\top \mathbf{X} \mathbf{X}^\top \mathbf{w}_x)(\mathbf{w}_y^\top \mathbf{Y} \mathbf{Y}^\top \mathbf{w}_y)}}
\end{equation}

\begin{figure}[t]
\centerline{\includegraphics[width=1\linewidth]{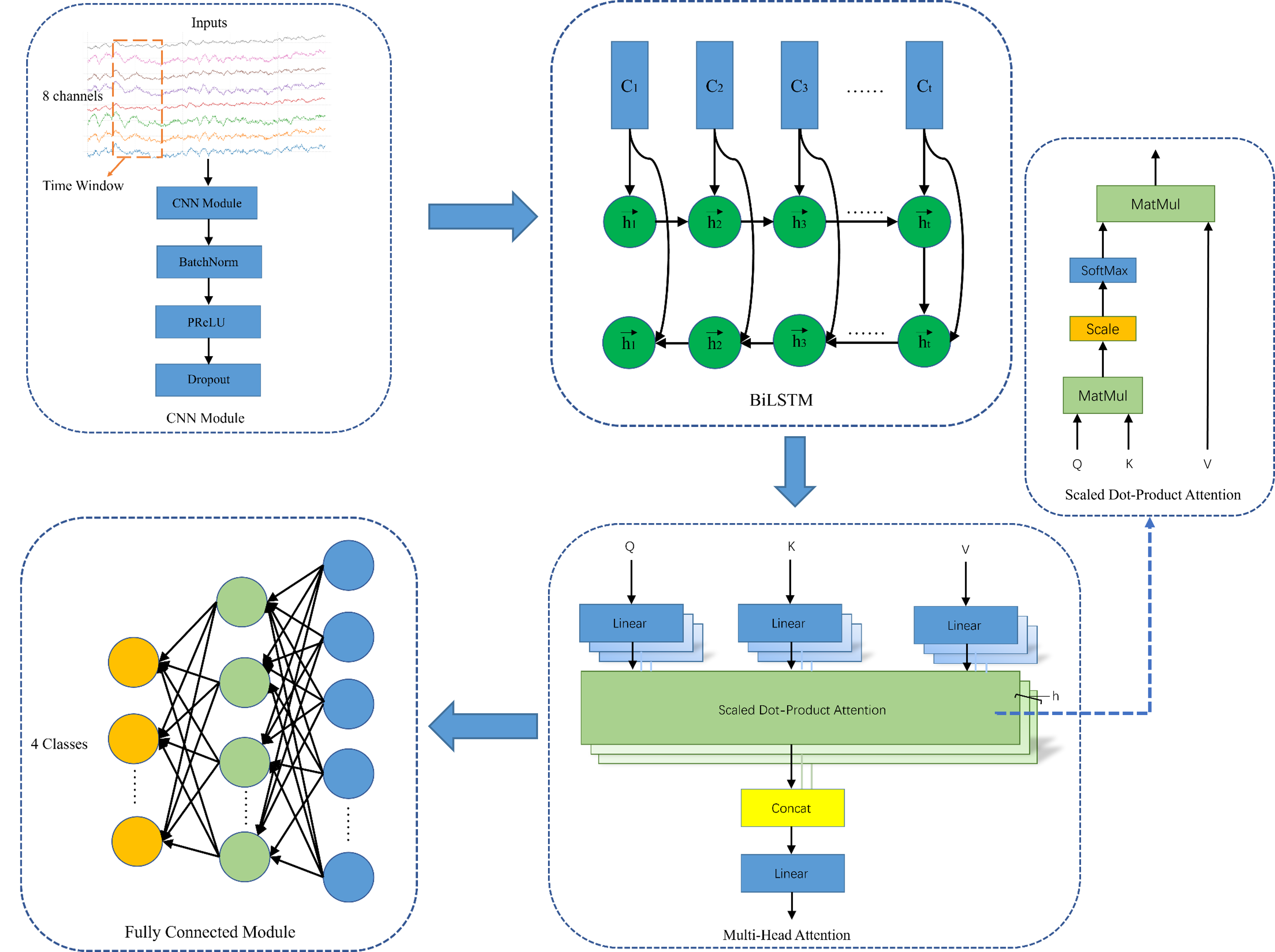}}
\caption{Structure of MACNN-BiLSTM.}
\label{fig4}
\end{figure}

\paragraph{Filter Bank Canonical Correlation Analysis} 

Filter Bank Canonical Correlation Analysis (FBCCA) is an extension of the classical CCA method designed to improve the robustness and accuracy of SSVEP classification~\cite{b21}. In FBCCA, the raw EEG signal $X$ is first decomposed into $N$ sub-band components $X_{SB1}, X_{SB2}, \dots, X_{SBN}$ using a set of band-pass filters that cover different frequency ranges. This decomposition aims to enhance sensitivity to the target frequency and its higher-order harmonics.

For each sub-band component, CCA is performed to calculate the maximal canonical correlation coefficient between the EEG sub-band signal and the reference signal $Y$:

\begin{equation}
\rho_k = \max_{a, b} \text{corr}(a^\top X_{SBk}, b^\top Y), \quad k = 1, 2, \dots, N
\end{equation}

The final decision score is then obtained by computing a weighted sum of the correlation coefficients from all sub-bands:

\begin{equation}
\rho_{\text{FBCCA}} = \sum_{k=1}^{N} w_k \cdot \rho_k
\end{equation}

where $w_k$ represents the weight assigned to the $k$-th sub-band, typically following an empirically defined decay function (e.g., $w_k = k^{-1.25} + 0.25$). By integrating frequency-specific information across multiple sub-bands, FBCCA significantly improves the robustness and classification performance, especially under low signal-to-noise ratio (SNR) conditions.

\paragraph{Convolutional Neural Networks} 
Convolutional Neural Networks (CNNs) have become a cornerstone in the field of deep learning, particularly for tasks involving spatially structured data such as images and multichannel time-series signals. CNNs leverage local connectivity and weight sharing through convolutional layers to effectively extract spatial hierarchies of features, making them highly suitable for applications in BCI systems \cite{b23}. In EEG-based classification, CNNs can automatically learn spatial and temporal dependencies from raw or minimally preprocessed signals, reducing the need for handcrafted features. 

\paragraph{Bidirectional Long Short-Term Memory Networks} 
Bidirectional Long Short-Term Memory Networks (BiLSTM) are an advanced form of recurrent neural networks (RNNs) that improve upon traditional LSTM networks by introducing a bidirectional structure\cite{b24}. Unlike standard LSTM, which processes data in a forward direction only, BiLSTM simultaneously processes sequences in both forward and backward directions as shown in Fig.~\ref{fig4}. This architecture enables the model to capture both past and future contextual dependencies in sequential data, making it especially suitable for complex time-series analysis such as EEG signals\cite{b25}.

\paragraph{Multihead Attention} 

Multi-head attention is a core component of the Transformer architecture, enabling the model to simultaneously focus on different positions of the input sequence. Unlike single-head attention, which computes attention once, the multi-head variant runs multiple self-attention mechanisms in parallel to capture richer semantic and positional relationships \cite{b26}.

As illustrated in Fig.~\ref{fig4}, each head uses the scaled dot-product attention mechanism, formulated as:

\begin{equation}
\text{Attention}(Q, K, V) = \text{softmax}\left(\frac{QK^T}{\sqrt{d_k}}\right)V
\end{equation}

where $Q$, $K$, and $V$ denote the query, key, and value matrices, and $d_k$ is the dimension of the key vectors. The scaling factor $\sqrt{d_k}$ helps to mitigate the risk of excessively large dot products when the dimensionality is high, which could lead to vanishing gradients or overly sharp softmax distributions.

In multi-head attention, outputs from multiple parallel attention heads are concatenated and projected to the final representation. This allows the model to jointly attend to information from different representation subspaces. In SSVEP decoding tasks, this mechanism enables more precise extraction of temporal and spatial patterns, thereby improving classification performance.

\subsection{SHAP Method}\
Explainable Artificial Intelligence (XAI) is a field dedicated to developing strategies for understanding the outcomes produced by machine learning (ML) algorithms\cite{b28}. It includes two primary types of interpretability approaches: post-hoc and intrinsic systems. Post-hoc systems provide localized explanations for individual model decisions, clarifying what the model predicts and why. An example of a post-hoc method is the LIME model. In contrast, intrinsic systems incorporate interpretability directly into their model structure, such as in linear regression\cite{b29}.

As none of the classifiers mentioned above is inherently explainable,it is necessary to employ post-hoc methods to explain the trained machine learning (ML) models without altering their structure. The goal is to understand how the model makes decisions. Impurity-based or permutation-based XAI algorithms (such as LIME and SHAP) achieve this by slightly modifying the input and measuring the resulting changes in predictions. These algorithms can be used to explain the model and understand its decision-making process without requiring any changes to the model itself.

SHAP is a model-agnostic interpretability framework grounded in cooperative game theory. It utilizes Shapley values to quantify the marginal contribution of each input feature to the model's prediction. These values are defined within an additive linear structure, where each feature is assigned an importance score that reflects its influence on a given prediction. A higher SHAP value indicates a greater impact of that feature on the model's output. By computing these contributions, SHAP enables a comprehensive and transparent explanation of the model's decision-making process.

\begin{figure}[b]
\centerline{\includegraphics[width=1\linewidth]{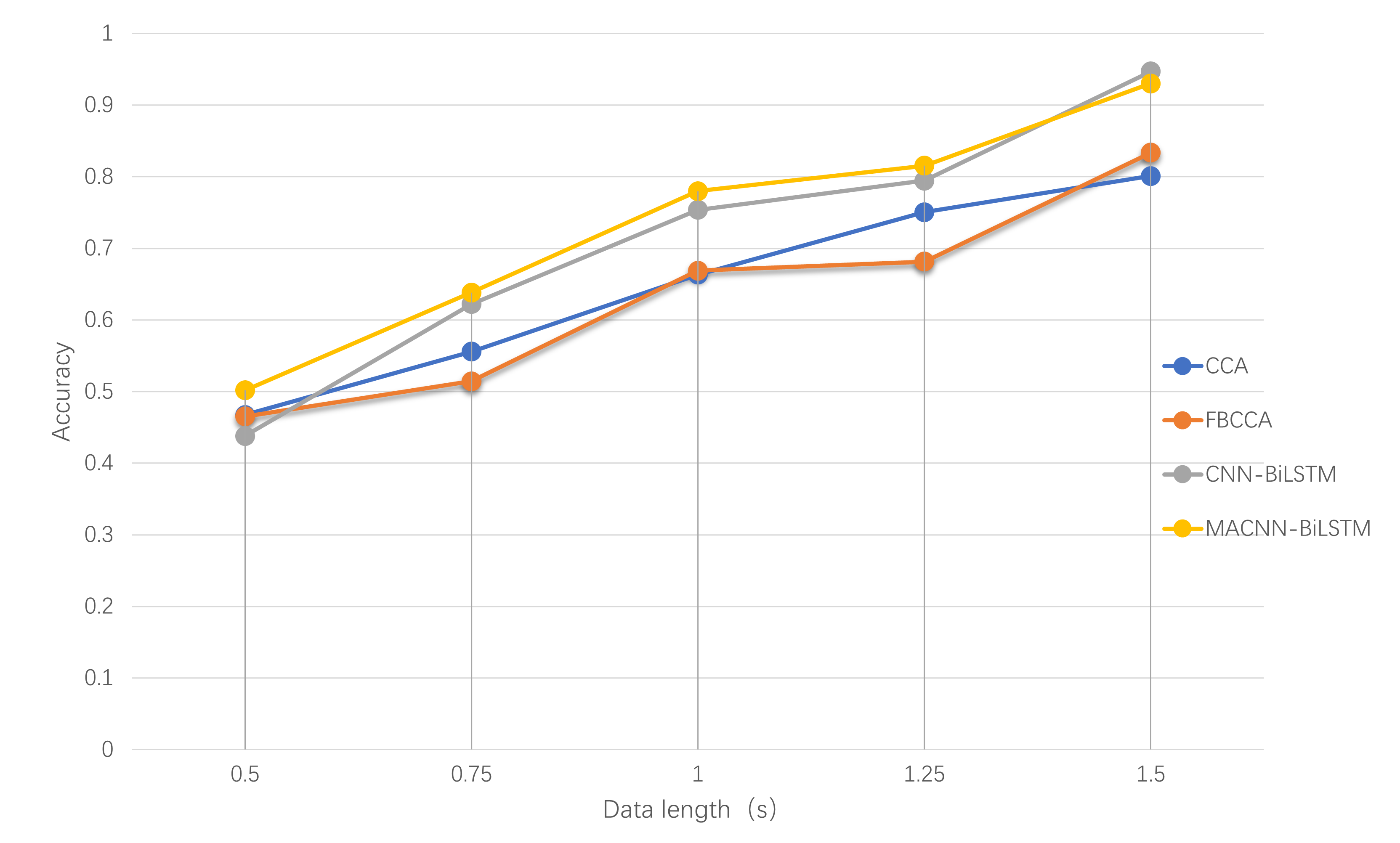}}
\caption{Average accuracy across subjects (0.5--1.5~s).}
\label{fig5}
\end{figure}

\section{Experiment and result analysis}
\subsection{Experimental Details}\
In this study, the model was trained using the PyTorch framework with the Adam optimizer. The initial learning rate was set to 0.01, and a weight decay of $1 \times 10^{-5}$ was applied to prevent overfitting. To further enhance training stability and generalization performance, a step learning rate scheduler (StepLR) was employed. This scheduler decays the learning rate by a factor of 0.9 every 100 epochs.

The training was conducted on a high-performance computing server equipped with 8 NVIDIA RTX A6000 GPUs, using Python 3.10 and CUDA 12.6.

For each trial, 7 seconds of EEG data were recorded, but only the middle 4 seconds were selected for training and evaluation. With a sampling rate of 1000 Hz, this results in 4000 time points per trial. Therefore, the input data for each subject's session was structured as a 3D tensor of size $40 \times 8 \times 4000$, where 40 represents the number of trials, 8 denotes the number of EEG channels, and 4000 corresponds to the number of sampled data points per trial.The ratio of training set to test set is 7:3. Effective EEG data are extracted by time windows of 0.5 s, 0.75 s, 1s, 1.25s and 1.5 s separately with the stepping time of 0.2 s.

\subsection{Recognition Performance}\
This study systematically evaluated the performance of AR-BCI system in motor intention recognition tasks, with a particular focus on the impact of data length and target identification algorithms on classification accuracy. We systematically compared four methods: CCA, FBCCA, CNN-BiLSTM, and the proposed MACNN-BiLSTM.

The within-subject recognition accuracy was evaluated for all seven subjects. The average accuracy across subjects under different time windows is shown in Table~\ref{tab1}. As illustrated in Fig.~\ref{fig5}, within the data length range of 0.5 to 1.5 seconds, the classification accuracy steadily increases as the data length extends. When the data length reaches 1.5 seconds, the proposed MACNN-BiLSTM algorithm achieves the highest average accuracy of 94.67\% across seven subjects, significantly outperforming the other algorithms: CCA (80.08\%), FBCCA (83.34\%), and CNN-BiLSTM (92.99\%).

\begin{figure}[t]
\centerline{\includegraphics[width=1\linewidth]{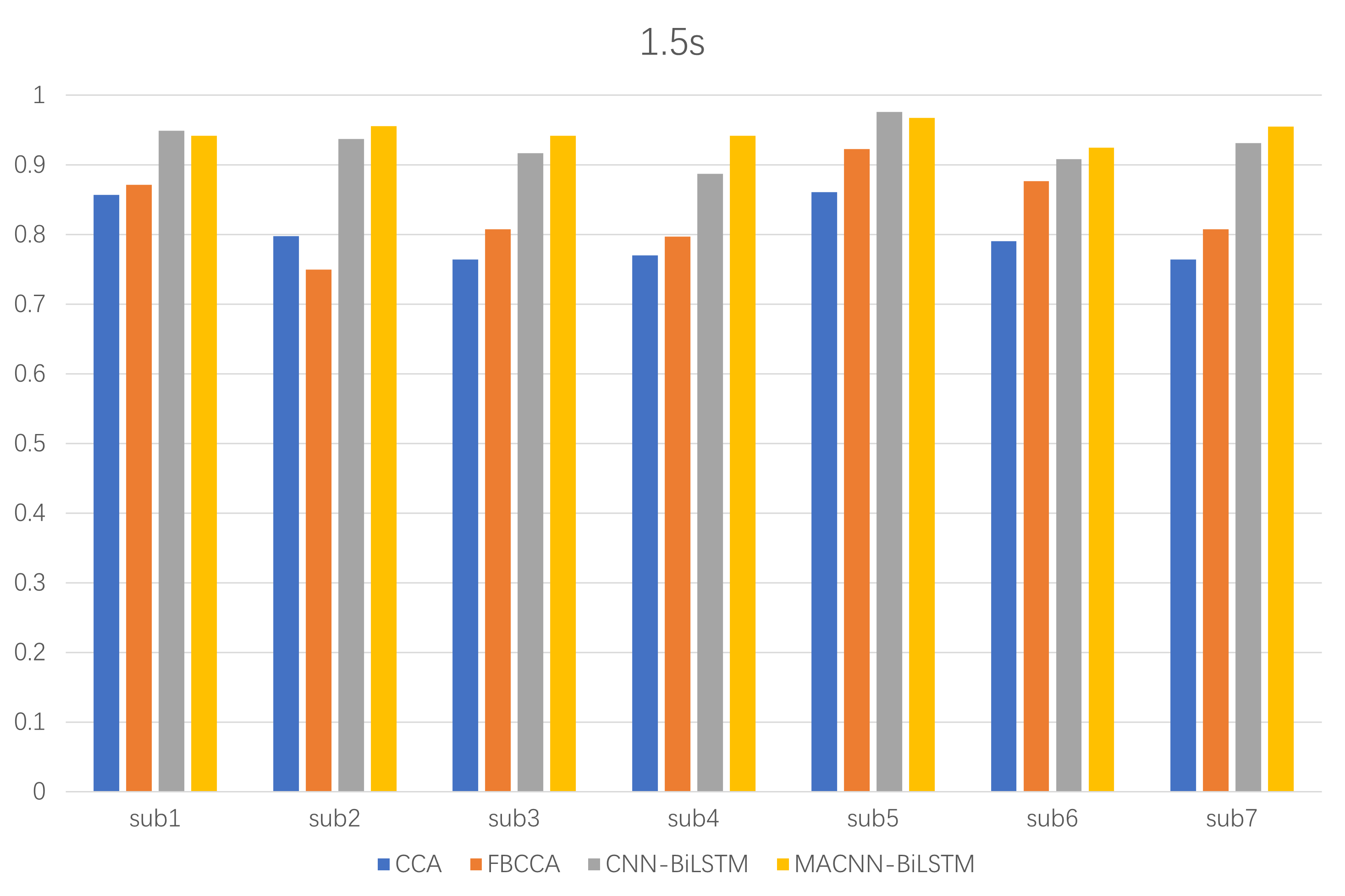}}
\caption{Recognition accuracy at the 1.5-second time window.}
\label{fig6}
\end{figure}

Fig.~\ref{fig5} presents the trend of average classification accuracy for all seven subjects across different data lengths, demonstrating the superior performance of the proposed AR-BCI system in SSVEP recognition. Fig.~\ref{fig6} shows the recognition accuracy at the 1.5-second time window. Notably, MACNN-BiLSTM maintains the highest recognition performance even under shorter time windows, highlighting its potential for rapid-response BCI applications.

\begin{table}[t]
  \centering
  \caption{Recognition accuracy of different methods across varying time windows}
  \begin{tabular}{lccccc}
    \toprule
    \textbf{Method} & \multicolumn{5}{c}{\textbf{Accuracy (\%)}} \\
    \cmidrule(lr){2-6}
                    & \textbf{0.5s} & \textbf{0.75s} & \textbf{1s} & \textbf{1.25s} & \textbf{1.5s} \\
    \midrule
    FBCCA           & 46.52 & 51.44 & 66.88 & 68.13 & 83.34 \\
    CCA             & 46.74 & 55.56 & 66.32 & 75.04 & 80.08 \\
    CNN-BiLSTM      & 43.80 & 62.21 & 75.36 & 79.44 & 92.99 \\
    MACNN-BiLSTM    & \textbf{50.18} & \textbf{63.82} & \textbf{77.97} & \textbf{81.53} & \textbf{94.67} \\
    \bottomrule
  \end{tabular}
  \label{tab1}
\end{table}

\begin{table}[t]
  \centering
  \caption{Descriptions of EEG features used in SHAP}
  \begin{tabular}{ll}
    \toprule
    \textbf{Feature Name} & \textbf{Description} \\
    \midrule
    peak\_freq & Peak frequency in the power spectral density (PSD)\\
    total\_psd & Total power across the full frequency range\\
    theta\_psd & Total power within the 4--8 Hz band \\
    alpha\_psd & Total power within the 8--12 Hz band\\
    beta\_psd & Total power within the 12--30 Hz band \\
    mean & Mean value of the signal \\
    std & Standard deviation of the signal\\
    skew & Skewness of the signal distribution\\
    max & Maximum amplitude value of the signal \\
    min & Minimum amplitude value of the signal \\
    \bottomrule
  \end{tabular}
  \label{tab2}
\end{table}

\subsection{Feature Importance Analysis}\

In this phase of the study, we applied the SHAP interpretability method to the best-performing MACNN-BiLSTM model to investigate the key EEG features the model relies on during motor intention recognition tasks. SHAP was used to interpret the model's prediction process by quantifying the contribution of each input feature to the final decision, thereby enhancing transparency and understanding of the model’s behavior.

\begin{figure}[t]
  \centering
  \subfigure[Start (6\,Hz)]{
    \includegraphics[width=0.22\textwidth]{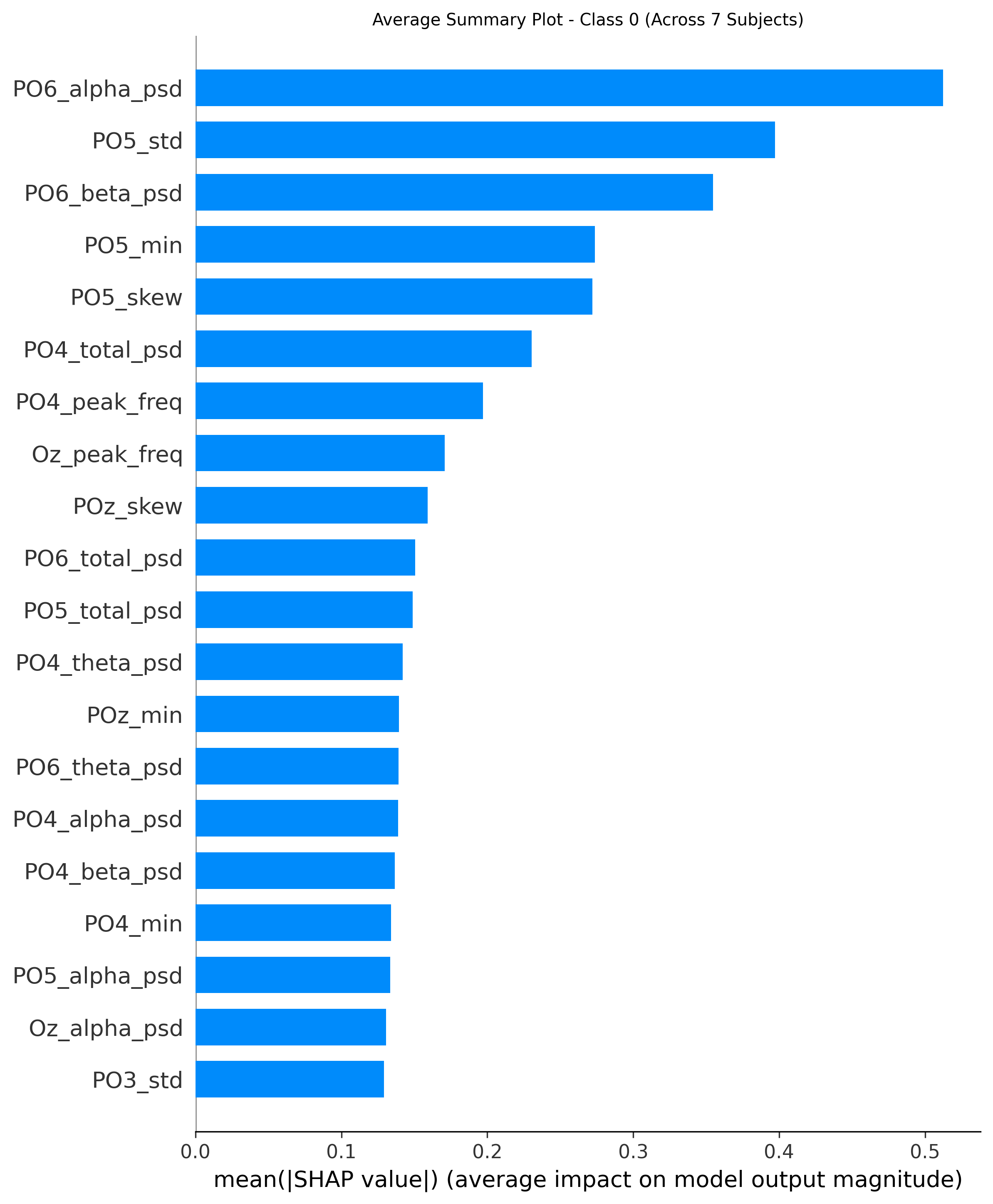}
  }
  \subfigure[Stop (8\,Hz)]{
    \includegraphics[width=0.22\textwidth]{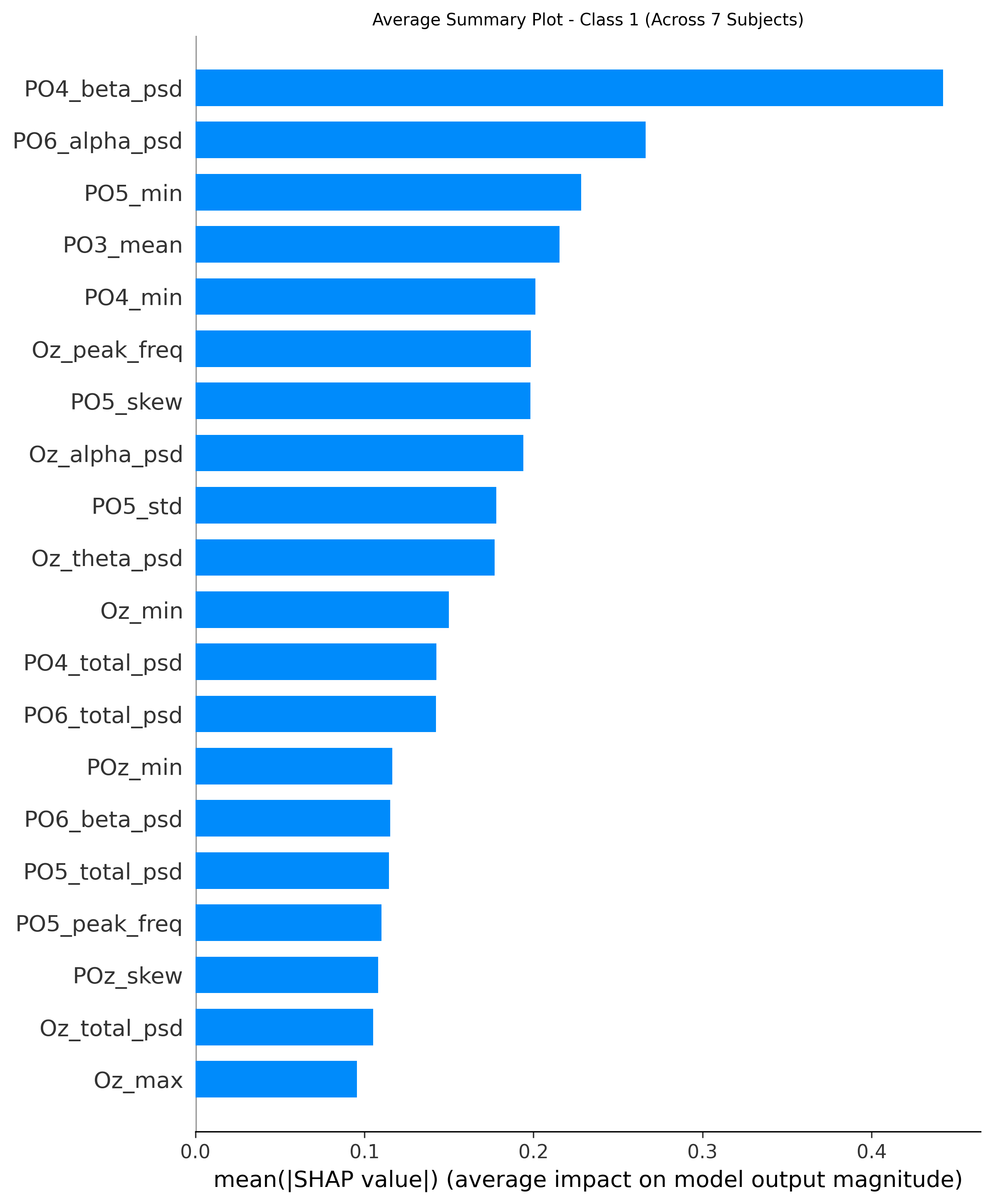}
  }
  \subfigure[Active (10\,Hz)]{
    \includegraphics[width=0.22\textwidth]{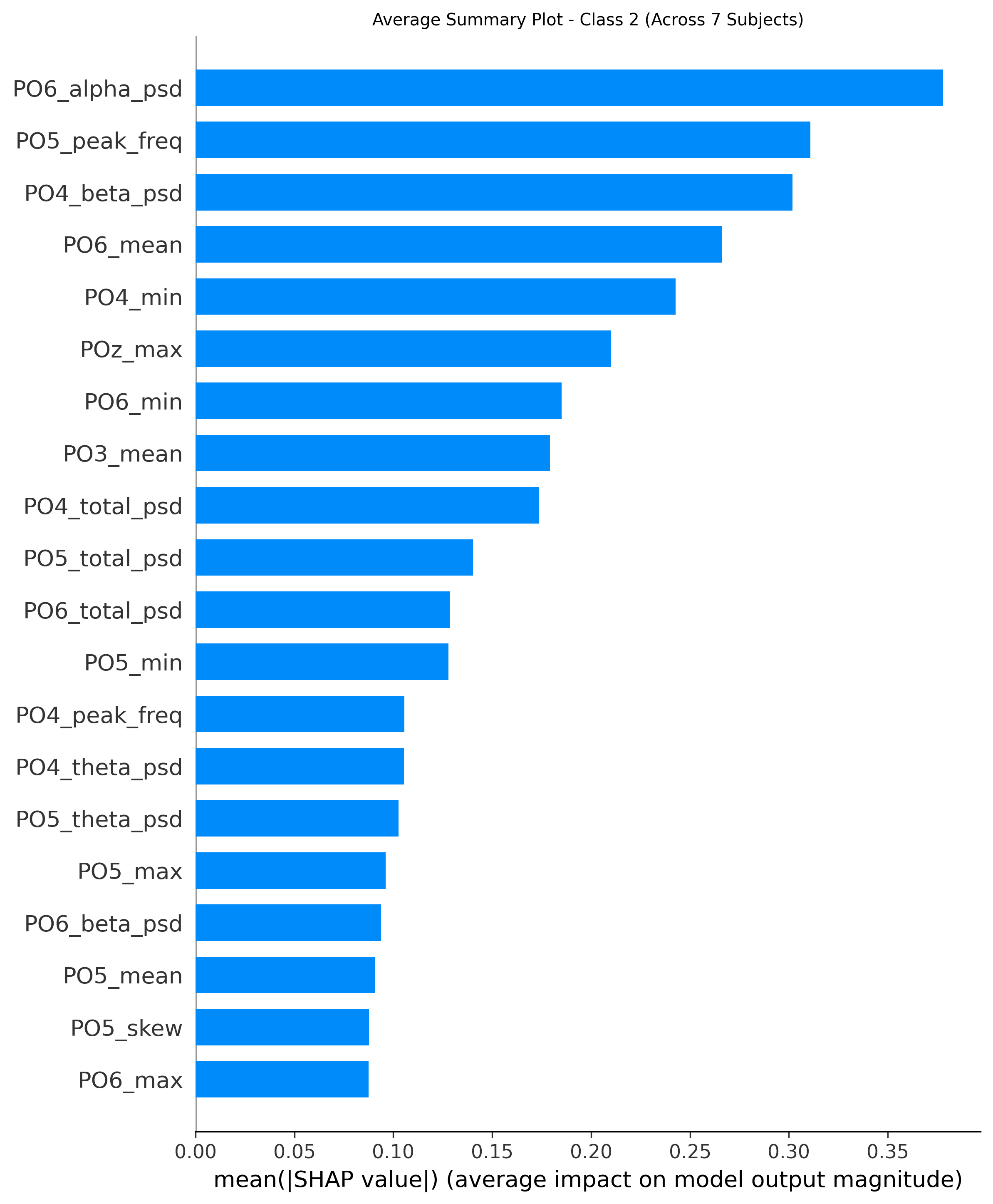}
  }
  \subfigure[Passive (12\,Hz)]{
    \includegraphics[width=0.22\textwidth]{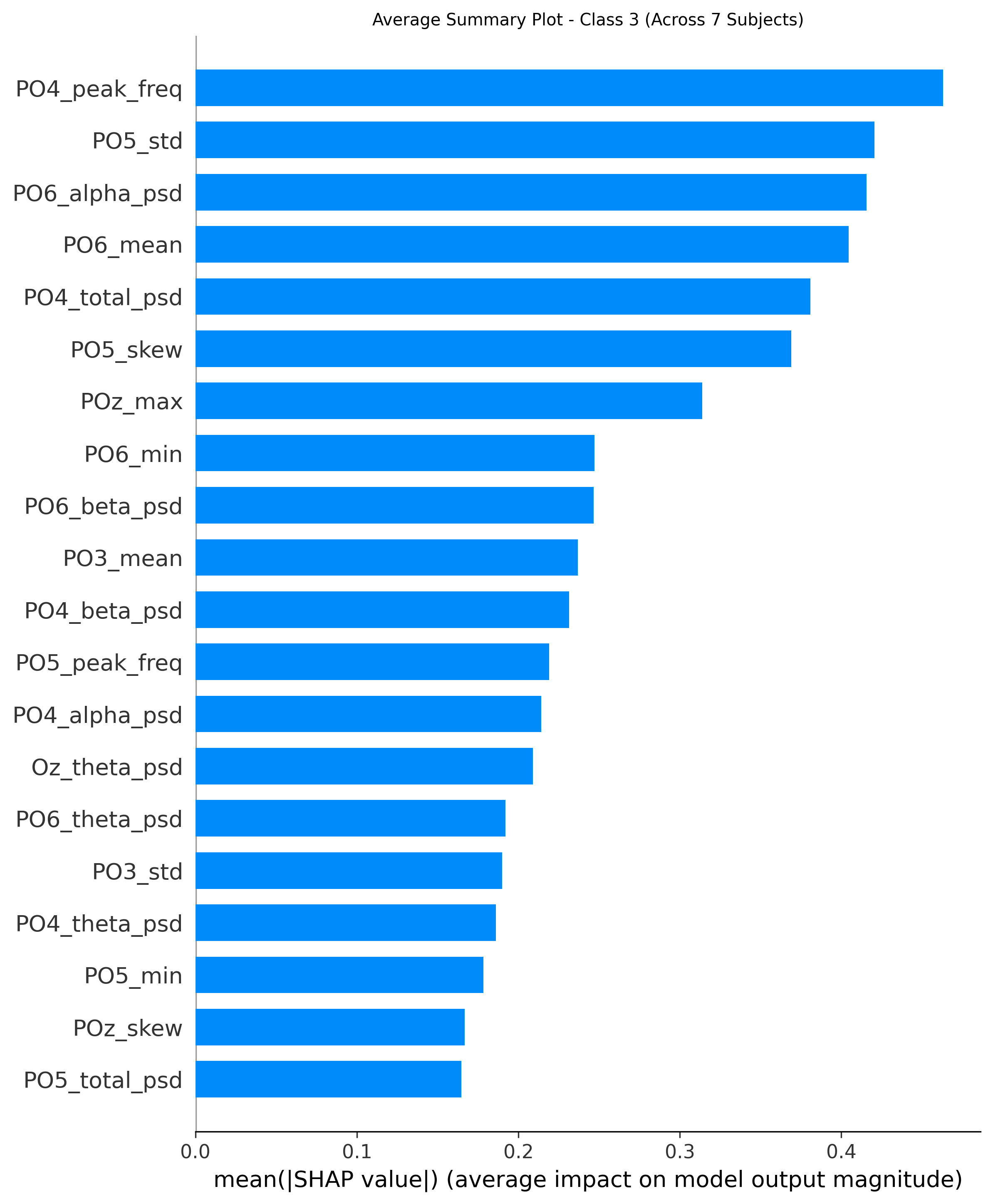}
  }
  \caption{SHAP feature importance across four classes.}
  \label{fig7}
\end{figure}

Table~\ref{tab2} summarizes the ten categories of EEG features used for SHAP analysis, covering both time-domain and frequency-domain indicators. Fig.~\ref{fig7} illustrates the SHAP value distributions across different EEG channels and their corresponding features, providing an intuitive visualization of feature importance in the classification task.

The classification labels, Class 0 to Class 3, correspond to four SSVEP stimulus frequencies, specifically representing the motor intention states of Start (6 Hz), Stop (8 Hz), Active (10 Hz), and Passive (12 Hz).

As shown in Fig.~\ref{fig7}, the SHAP summary plot presents the contribution of each EEG feature across different channels and frequency domains.By aggregating SHAP values across all four classes, we identified PO6 Alpha PSD, PO5 Std, and PO4 Beta PSD as the three most influential features in the model's prediction process. These features reflect the critical role of frequency-specific and statistical attributes, particularly in the occipital region, which is closely associated with visual and cognitive processing. The results highlight the effectiveness of combining spatial, spectral, and statistical features in SSVEP-based intention recognition.

\section{Conclusion}

This study presents an AR-SSVEP-based BCI system that integrates augmented reality with advanced neural signal processing techniques for motor intention recognition. The proposed MACNN-BiLSTM model achieved a classification accuracy of 94.67\% at a data length of 1.5 seconds, outperforming conventional approaches. SHAP-based interpretability analysis identified PO6 Alpha PSD, PO5 Std, and PO4 Beta PSD as the most discriminative EEG features contributing to the model’s decisions.

The integration of AR visualization, deep learning, and explainable AI demonstrates promising potential for neurorehabilitation applications. The system achieves low-latency recognition and interpretable decision-making, addressing two critical challenges in practical BCI deployment—responsiveness and transparency. Future work will focus on clinical validation with patient populations and enhancing cross-subject generalization to support broader real-world rehabilitation scenarios.

\end {document}